\documentclass[10pt,twocolumn,letterpaper]{article}

\usepackage{cvpr}              

\usepackage[dvipsnames]{xcolor}

\definecolor{cvprblue}{rgb}{0.21,0.49,0.74}
\usepackage[pagebackref,breaklinks,colorlinks,citecolor=cvprblue]{hyperref}

\def\paperID{367} 
\def\confName{3DV\xspace}
\def\confYear{2025\xspace}

\usepackage{svg}
\usepackage[utf8]{inputenc} 
\usepackage[T1]{fontenc}    
\usepackage{hyperref}       
\usepackage{url}            
\usepackage{booktabs}       
\usepackage{amsfonts}       
\usepackage{nicefrac}       
\usepackage{microtype}      
\usepackage{xcolor}         
\usepackage{caption}

\usepackage{graphicx}
\usepackage{mathtools}
\usepackage{amssymb}
\usepackage{booktabs}
\usepackage{bbm}
\usepackage{wrapfig}
\usepackage{amsmath, bm}
\usepackage{algorithm, algpseudocode}
\usepackage{algorithmicx}
\usepackage{caption}
\newcommand{\comment}[1]{}

\newcommand {\yk}[1]{{\color{gray}\textbf{YK: }#1}\normalfont}

\newcommand{\subsecref}[1]{Section~\ref{sec:#1}}

\title{Pixel-Aligned Multi-View Generation with Depth Guided Decoder}

\author{Zhenggang Tang,$^{1}$ Peiye Zhuang,$^{2}$ Chaoyang Wang,$^{2}$ Aliaksandr Siarohin,$^{2}$ Yash Kant$^{3}$ \\
Alexander Schwing,$^{1}$ Sergey Tulyakov$^{2}$, Hsin-Ying Lee$^{2}$ \\
$^{1}$University of Illinois Urbana-Champaign $^{2}$Snap Inc. $^{3}$University of Toronto. }

\begin{document}

\maketitle

\begin{abstract}
  
The task of image-to-multi-view generation refers to generating novel views of an instance from a single image. Recent methods achieve this by extending text-to-image latent diffusion models to multi-view version, which contains an VAE image encoder and a U-Net diffusion model. Specifically, these generation methods usually fix VAE and finetune the U-Net only. However, the significant downscaling of the latent vectors computed from the input images and independent decoding leads to notable pixel-level misalignment across multiple views.
To address this, we propose a novel method for pixel-level image-to-multi-view generation. Unlike prior work, we incorporate attention layers across multi-view images in the VAE decoder of a latent video diffusion model. Specifically, we introduce a depth-truncated epipolar attention, enabling the model to focus on spatially adjacent regions while remaining memory efficient. Applying depth-truncated attn is challenging during inference as the ground-truth depth is usually difficult to obtain and pre-trained depth estimation models is hard to provide accurate depth. Thus, to enhance the generalization to inaccurate depth when ground truth depth is missing, we perturb depth inputs during training. During inference, we employ a rapid multi-view to 3D reconstruction approach, NeuS, to obtain coarse depth for the depth-truncated epipolar attention. Our model enables better pixel alignment across multi-view images. Moreover, we demonstrate the efficacy of our approach in improving downstream multi-view to 3D reconstruction tasks.

\end{abstract}

\begin{figure*}[th]
\centering
\includegraphics[width=1.0\linewidth]{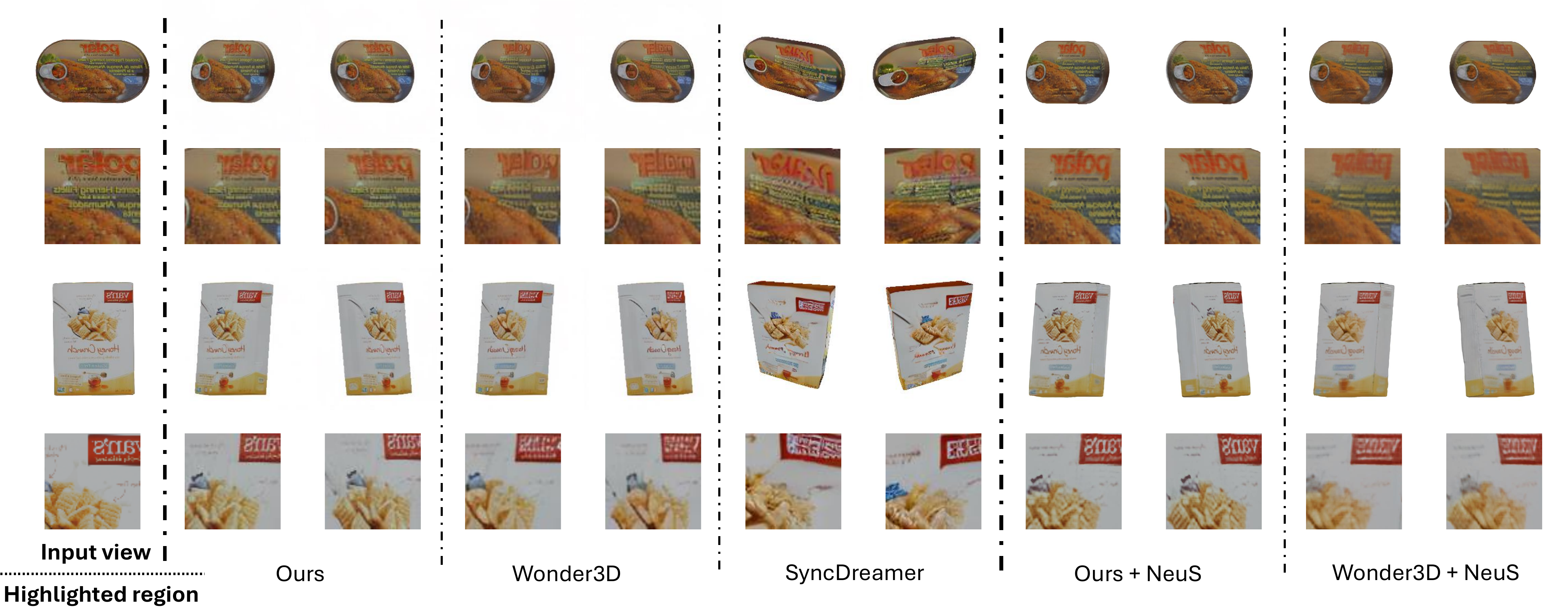}
\captionof{figure}{
    \textbf{Visualization of our method}. Comparing to the baseline methods (column 4-7, 10-11), our proposed method enables to generate pixel-aligned multi-view images, which can lead to better 3D reconstruction quality.
}
\label{fig:teaser}
\end{figure*}

\section{Introduction}
Multi-view images that show an object from a small number of different viewpoints, have emerged as a commonly used auxiliary representation in 3D generation. 
To obtain multi-view images, multi-view diffusion models~\cite{zero123,mvdream,syncdreamer} are fine-tuned from a large-scale 2D diffusion model~\cite{ldm} and inherit their generalizability since the model is trained on huge data. 
The resulting multi-view diffusion models then serve downstream tasks, e.g.,  a subsequent 3D reconstruction in a 3D generation pipeline~\cite{instant3d,dmv3d,lgm}, or act as approximate 3D priors for distillation-based 3D~\cite{magic123,consistent123} or 4D generation~\cite{4dfy,aligngaussian,4dgen}.

Despite promising results, it is challenging for current multi-view generation methods to achieve pixel-level image alignment across views. 
The coarse alignment achieved with current methods introduces ambiguity in subsequently employed reconstruction methods. as shown in Fig.~\ref{fig:teaser}, irrespective of whether per-instance optimization~\cite{neus} or feed-forward methods~\cite{instant3d} are used for 3D generation, they get blurred results due to pixel-level misalignment. 
Pixel-level alignment issues arise because existing multi-view diffusion models are mostly fine-tuned from an image diffusion model with additional multi-view attention~\cite{wonder3d,mvdream} or an intermediate implicit 3D representation~\cite{syncdreamer,dmv3d}. 
Notably, the diffusion process occurs in a latent space with limited resolution, and decoding is performed independently for each frame without cross-view communication, making pixel-level alignment difficult. 
To improve, some multi-view diffusion models are fine-tuned from video diffusion models with camera trajectory control~\cite{sv3d,v3d}. 
Although the multi-view latents are jointly decoded using a video decoder, achieving pixel-level alignment remains challenging due to the sparsity of adjacent multi-view frames.

In this work, we propose to address the pixel-level alignment issue by improving existing VAE decoders. 
Following prior multi-view generation works, We adopt the VAE decoder from Stable Video Diffusion~\cite{svd} as our backbone. Differently, to enable cross-view attention at higher-resolution and and achieve better pixel-level multiview alignment, we modify the VAE decoder in two ways:
First, we propose a depth-truncated epipolar attention mechanism applied to high-resolution layers. This attention mechanism extracts cross view features that are crucial for better feature alignment. 
Different from conventional epipolar attention, the depth-truncated epipolar attention not only helps models focus on critical regions, but also enables information aggregation at high resolution. 
However, the depth information is not available during inference. 
Moreover, the multi-view latents are often not accurately aligned.
Second, to solve this, we augment data with structured-noise depth to mitigate the domain gap between training and inference. We propose to augment data with structured-noise depth, appending both high- and low-frequency noise to the ground-truth depth. 
Then during inference, we simply employ depth predicted by an off-the-shelf multi-view 3D reconstruction method~\cite{neus}. This is feasible as we obtain a model that is more robust to imperfect predictions.

We conduct extensive qualitative and quantitative experiments against baseline methods. 
We visually compare with other multi-view generation methods by adopting the same 3D reconstruction methods~\cite{neus} and quantitatively measure PSNR, SSIM, LPIPS, and the number of correspondences,  on the reconstructed 3D objects. 
The proposed method performs favorably against existing state-of-the-art multi-view generation methods.

\section{Related work}

\paragraph{3D generation.}
Conventional 3D generative models are train on 3D data and have employed various  representations, including point clouds~\cite{pointe,achlioptas2018learning}, voxels~\cite{infinicity,smith2017improved}, meshes~\cite{zhang2021sketch2model}, implicit functions~\cite{sdfusion,shape,cheng2022cross,autosdf}, etc.
However, the scarcity of 3D data limits the quality and diversity of these methods. 
In contrast, image diffusion models have witnessed superior quality and generalizability due to available large-scale data.
To utilize pre-trained image diffusion models for 3D generation, Score Distillation Sampling (SDS)~\cite{dreamfusion} and its variants~\citep{magic3d,sjc,fantasia3d,prolificdreamer,hifa} have been proposed to distill knowledge from 2D models in a per-instance optimization manner, taking minutes to hours for each generation. 
Recently, to circumvent time-consuming optimization, feed-forward methods~\cite{instant3d,lrm,lgm,dmv3d} have emerged. They use multi-view images as an auxiliary representation followed by 3D reconstruction. 
In this work, we focus on  pixel-aligned multi-view image generation to facilitate better 3D reconstruction, ultimately leading to better 3D generation. 

\comment{
\begin{enumerate}
    \item 3D diffusion like SDFusion, XCube.
    \item SDS based methods like LucidDream.
    \item feed-foward methods given MV generation like instant3D, LGM.
\end{enumerate}
}

\paragraph{Multi-view image generation.}
To inherit the generalizability, multi-view diffusion models are mostly fine-tuned from large-scale image diffusion models~\cite{ldm} using synthetic 3D object data~\cite{objaverse}. 
To adapt from image diffusion models, multi-view diffusion models incorporate multi-view cross attention~\cite{zero123++, wonder3d} or adopt intermediate 3D representations like voxels~\cite{syncdreamer} or a triplane-based neural radiance field (NeRF)~\cite{dmv3d}.
With recent advances in video diffusion models, some methods propose to fine-tune from video diffusion models with camera trajectory control~\cite{sv3d} or by treating multi-view generation as a form of image-to-video translation.
However, irrespective of the approach, existing efforts still struggle to synthesize pixel-level aligned multi-view images. 

\comment{
\begin{enumerate}
    \item multi-view 2D diffusion like Wonder3D, MVDream.
    \item multi-view diffusion with 3D representation: SyncDreamer, DMV3D
    \item video-diffusion finetuned on multi-view settings like SVD, V3D, SV3D.
\end{enumerate}
Note that we do not work on 3D generation, but we are improving its upstream task: MV generation and can improve 3D gen subsequently.}

\paragraph{Epipolar attention in multi-view stereo and multi-view generation.}
Multi-view Stereo (MVS) is a classic task aiming to reconstruct  3D scenes from multiple  views that are assumed to be given. 
With the advent of deep learning, learning-based MVS methods~\cite{gu2020cascade,eppmvsnet,wang2021patchmatchnet,wei2021aa,yi2020pyramid} have dominated the field, often improving upon traditional approaches~\cite{tola2012efficient,schonberger2016structure,galliani2015massively,campbell2008using}.
Learning-based methods formulate cost volumes by incorporating 2D semantics and 3D spatial associations. 
Our work is related to a stream of methods~\cite{mvster,eppmvsnet,mvs2d,mvsformer} employing epipolar attention to help aggregate information from multi-view images.  
Epipolar attention mechanisms help in aligning and combining multi-view features more effectively, enhancing the accuracy and consistency of the reconstructed 3D scenes. 

\comment{
\begin{enumerate}
    \item epipolar attention on MVS.
\end{enumerate}
}
 \begin{figure*}[t]
 \centering
  \includegraphics[width=1\textwidth]{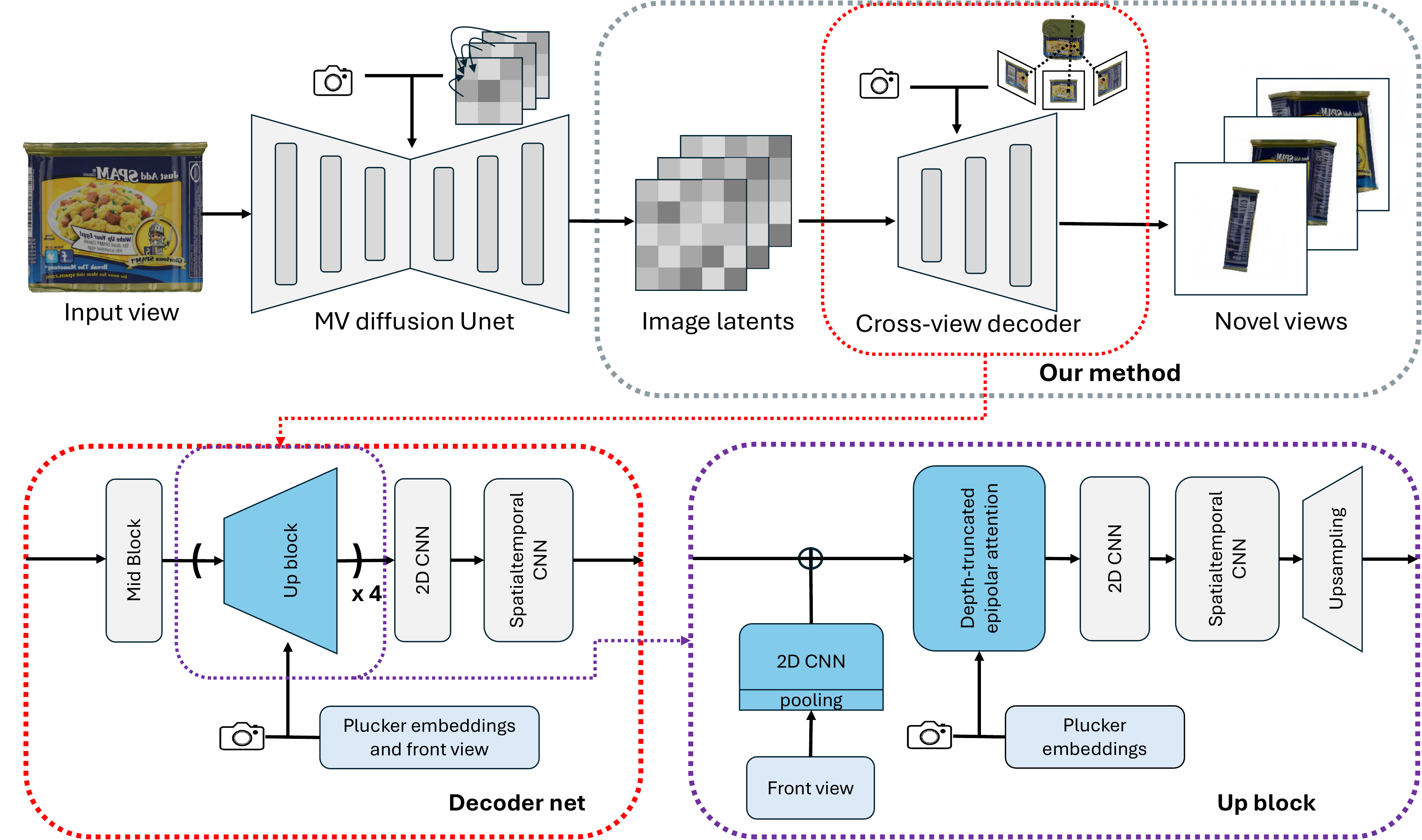}
 \caption{
 \textbf{Overview.}
 (top) We aim to achieve pixel-aligned multi-view image generation from the multi-view latents, either encoded or generated by a multi-view diffusion model. For this, we focus on improving the decoder. 
 (bottom-left) The decoder contains four Up-blocks to upsample the resolution from $32$ to $256$. 
 (bottom-right) We propose several additions, highlighted with blue color. 
 We add a condition from the input front-view image, and a depth-truncated epipolar attention mechanism. 
 Note that the $4^\text{th}$ Up-block does not have an upsampling layer, as the resolution is not changed.}
 \label{fig:pip}
 \end{figure*}

\section{Method}
We aim to generate  multi-view images with better pixel-level alignment. 
For this, we focus on improving the decoder of a latent diffusion model. 
Specifically, the proposed method is based on the decoder from Stable Video Diffusion (SVD)~\cite{svd}. 

To improve pixel-level alignment, we propose a depth-truncated epipolar attention mechanism. It aggregates features from multi-view latents by making use of depth information. 
To further mitigate the domain gap between the ground-truth depth used in training and the predicted depth used in inference, we propose a structured-noise depth augmentation strategy.
The strategy can also help handle the imperfect generated multi-view latents during inference. 

In the following, we first provide an overview of the proposed approach in \subsecref{overview}. 
We then introduce the depth truncated epipolar attention mechanism in \subsecref{epipolar}, the structured-noise depth augmentation  strategy in \subsecref{noisy}, and the implementation details in \subsecref{imp}.

\subsection{Overview}
\label{sec:overview}

To generate pixel-level aligned multi-view images, we propose a cross-view decoder with depth-truncated epipolar attention (Fig.~\ref{fig:pip}). The decoder takes image latents as input and outputs multi-view RGBs. Latents are generated from an image-conditioned multi-view diffusion model, which typically uses multi-view self-attention to learn low-resolution consistency. In contrast, we propose a truncated epipolar attention module which is employed in each Up-block of the decoder. Moreover, we find that  additionally conditioning the decoder on the available front view improves the results.

\begin{figure}[t]
    \centering
    \subfloat[Full epipolar attention.]{%
        \includegraphics[width=0.4\textwidth]{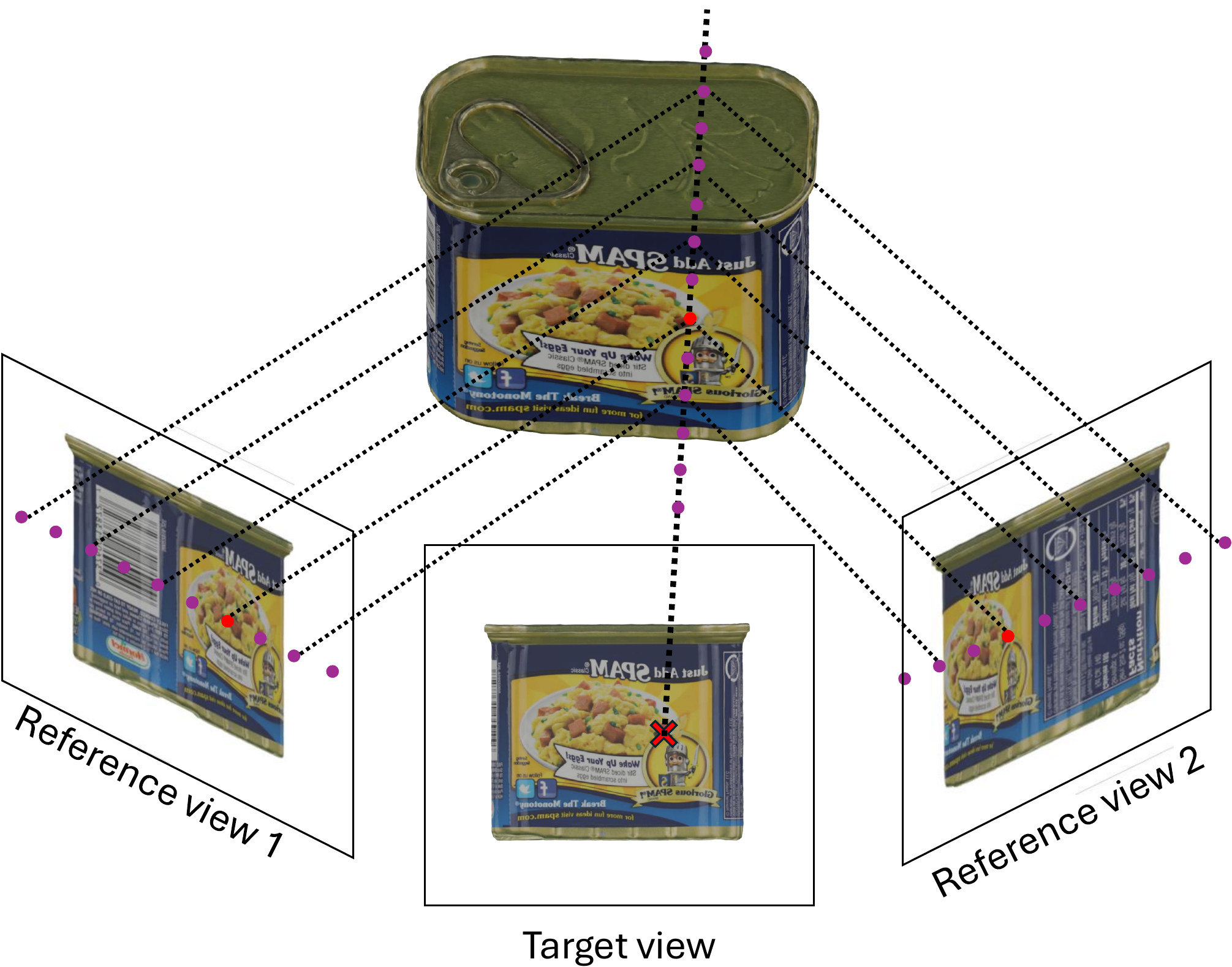}
        \label{fig:epi}
    }
    \hfill
    \subfloat[Depth-truncated epiplar attention.]{%
        \includegraphics[width=0.4\textwidth]{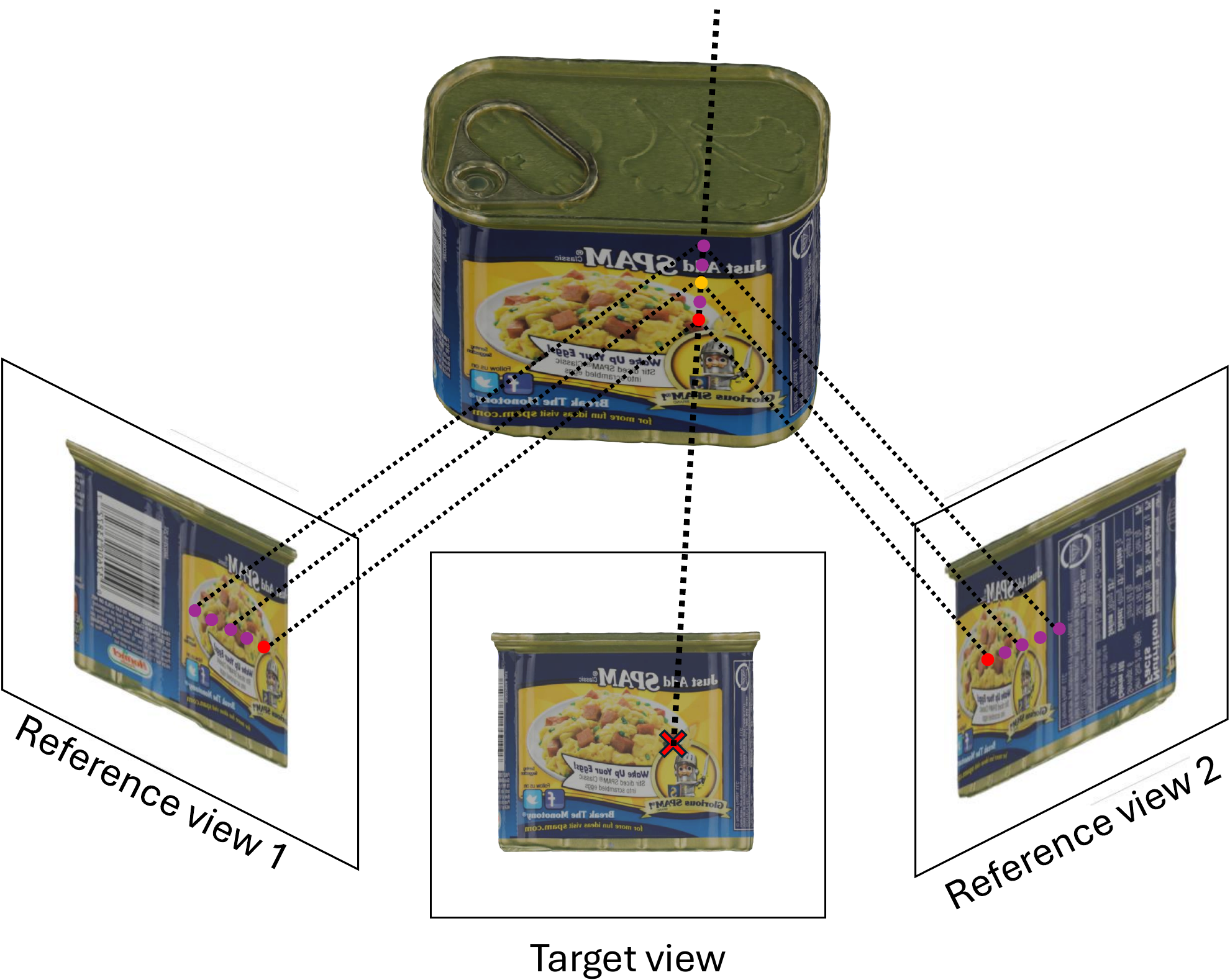}
        \label{fig:epit}
    }
    \caption{\textbf{Epipolar Attention.}
    (a) Full epipolar attention aggregates information along the whole epiploar line,  covering unnecessary ranges (only the red dot is the correct position), which limits applicability to lower resolution representations due to memory constraints. 
    (b) Depth-truncated epipolar attention samples only points near the 3D location of that pixel (the red dot).
    It enables epipolar attention on higher-resolution representations and  improves information aggregation. 
    }
    \vspace{-3mm}
    \label{fig:combined_fig}
\end{figure}

\subsection{Depth-truncated epipolar attention 
}
\label{sec:epipolar}
To generate pixel-level aligned multi-view images, the decoder needs to gather and process information from multi-view latents.
An epipolar attention mechanism is an excellent candidate for this task, because it permits to combine information from corresponding points across views. 
Importantly, to attain a more accurate pixel-level information exchange, the attention is preferably applied to any resolution, particularly also on higher resolution latents. 
However, a vanilla epipolar attention mechanism often spreads too much attention on irrelevant parts, which makes it difficult for the network  to learn to extract the correct adjacent features. Moreover, it also consumes a lot of memory and  easily leads to out-of-memory errors given current hardware memory constraints, even on high-end equipment. 
To address this, we propose a depth-truncated epipolar attention mechanism.
This approach not only aggregates multi-view information at higher resolutions, but also further improves the quality by enabling the model to focus on crucial regions.

Concretely, consider a feature map from a referenced view $F^{\mathbf{ref}}$, and feature maps from $N_v$ other views $\{F^j\}_{j\neq \mathbf{ref}}$. 
For a source point $s$ on $F^{\mathbf{ref}}$, we can get the epipolar lines $\{l^j\}_{j\neq \mathbf{ref}}$ on the other views.
Instead of using all points on the epipolar lines, we only sample points around the regions of interest. 
Specifically, given a known or estimated depth value $z$ and the camera intrinsics, we can unproject the point to 3D space $s_\mathbf{3D}$.
We the sample $N_p$ points $\{p_i\}$ around $s_\mathbf{3D}$ in range $[-r,r]$ in a stratified way along the 3D line formed by $s_\mathbf{3D}$ and  $T_\mathbf{c2w}\cdot s$, where $T_\mathbf{c2w}$ is a camera-to-world transformation. 
We then project these points to the epiplor lines $\{l^j\}$ on the other views to extract features.

To compute the cross attention among views, we first aggregate features across views. 
For the $N_p$ sampled points, we get features $\{f_i^j\}_{i=1,\ldots,N_p, \; j=1,\ldots,N_v}$ after projecting the points onto the  epipoloar line on the $j^\text{th}$ feature map. 
For each point, we aggregate these features by a concatenation operation followed by a 2-layer MLP, 
\begin{equation}
\begin{aligned}
    & f^\mathbf{mv}_i=\mathbf{MLP}(\mathbf{concat}(\{f_i^{j}\}_j)),\\
    & \mathbf{MLP}: \mathbb{R}^{N \times \text{dim}} \rightarrow \mathbb{R}^{\text{dim}}
, f^\mathbf{mv}_i \in \mathbb{R}^d.
\end{aligned}
\end{equation}
Then, we aggregate the features across $N_p$ points by stacking along the feature dimension and get $F^\mathbf{mv} \in \mathbb{R}^{N_p \times d}$.

We use this feature map to compute the keys and values of classic attention while the queries are computed from the reference view, i.e., 
\begin{equation}
\begin{aligned}
Q &= W_Q \cdot F^{\mathbf{ref}}, W_Q \in \mathbb{R}^{d \times HW}\\
K &= W_K \cdot F^\mathbf{mv}, W_K \in \mathbb{R}^{d \times N_p}\\
V &= W_V \cdot F^\mathbf{mv}, W_V \in \mathbb{R}^{d \times N_p}\\
\end{aligned}
\end{equation}

We apply the depth-truncated epipolar attention on all latent resolutions (from 32 to 256) in all Up-blocks of the decoder.

\comment{
\begin{enumerate}
\item attention overview: for each pixel in the reference feature map i (1 to 6) in different hierarchies (resolutions). given the depth value $z$ of that pixel and orthogonal camera intrinsics, we unproject it to 3D and sample $N_p$ points in a stratified way near $z$, then project back to other views to query features. Then there should be two information aggregation. one cross views, another cross points.

\item cross view: features of 6 views concatenated together, then 2-layer MLP. shape: $[N_v, d] \rightarrow [N_v * d] \rightarrow$ through MLP $\rightarrow [d]$

\item attention cross $N_p$ points, Q are mapped from features of the reference view, K and V are mapped from the feature aggregated above.
\end{enumerate}
}


\subsection{Structured-noise depth augmentation}
\label{sec:noisy}
The proposed depth-truncated epipolar attention mechanism requires access to depth information. 
During training, we leverage 3D data to obtain ground-truth depth. 
During inference, we can predict depth using off-the-shelf depth predictors. 
However, the predicted depth is usually imperfect. 
Furthermore, the multi-view latents are encoded from ground-truth 3D assets during training, but are generated by the diffusion process during inference. 
That is, the multi-view latents we are decoding might not be accurately aligned. 
Therefore, we need a strategy to mitigate the domain gap. 

Rather than warping the ground truth latents to simulate the misalignment of generated latents, we warp the ground truth depth, which can be regarded as equivalently warping the latents. For this, we propose structured-noise depth augmentation as the noising process.
During training, we uniformly sample noise in lower resolution $({3,64,128})$ hierarchically. 
We then upsample these noises to the $256$ resolution, and add them to the $256$ resolution ground-truth depth map $D$.
We formulate the process as follows:
\begin{equation}
\begin{aligned}
Z_i &\sim \mathcal{U}(-s_i, s_i)^{i \times i}, i\in\{3, 64, 128\} \\
\{ Z_i'\} &= \mathbf{Upsample}(\{Z_i\}, 256), Z_i'\in \mathbb{R}^{256 \times 256} \\ 
D' &= D + Z_3' + Z'_{64} + Z'_{128}
\end{aligned}
\end{equation}
The noisy depths $D'$ now contain both high and low frequency noise. Note that the depth map will be pooled to different resolution for different hierarchies of Up-blocks. Note, compared to the naive strategy which perturbs each depth pixel  with  independent Gaussian or uniform noise, our strategy does not cause the noise to be cancelled out in lower resolutions. This makes our method more robust.

During inference, we simply use the predicted depth (we use Neus~\cite{neus} in this work), as we find our model to be robust to inaccurate predictions. 

\begin{figure*}[p]
\centering
\includegraphics[width=0.94\textwidth]{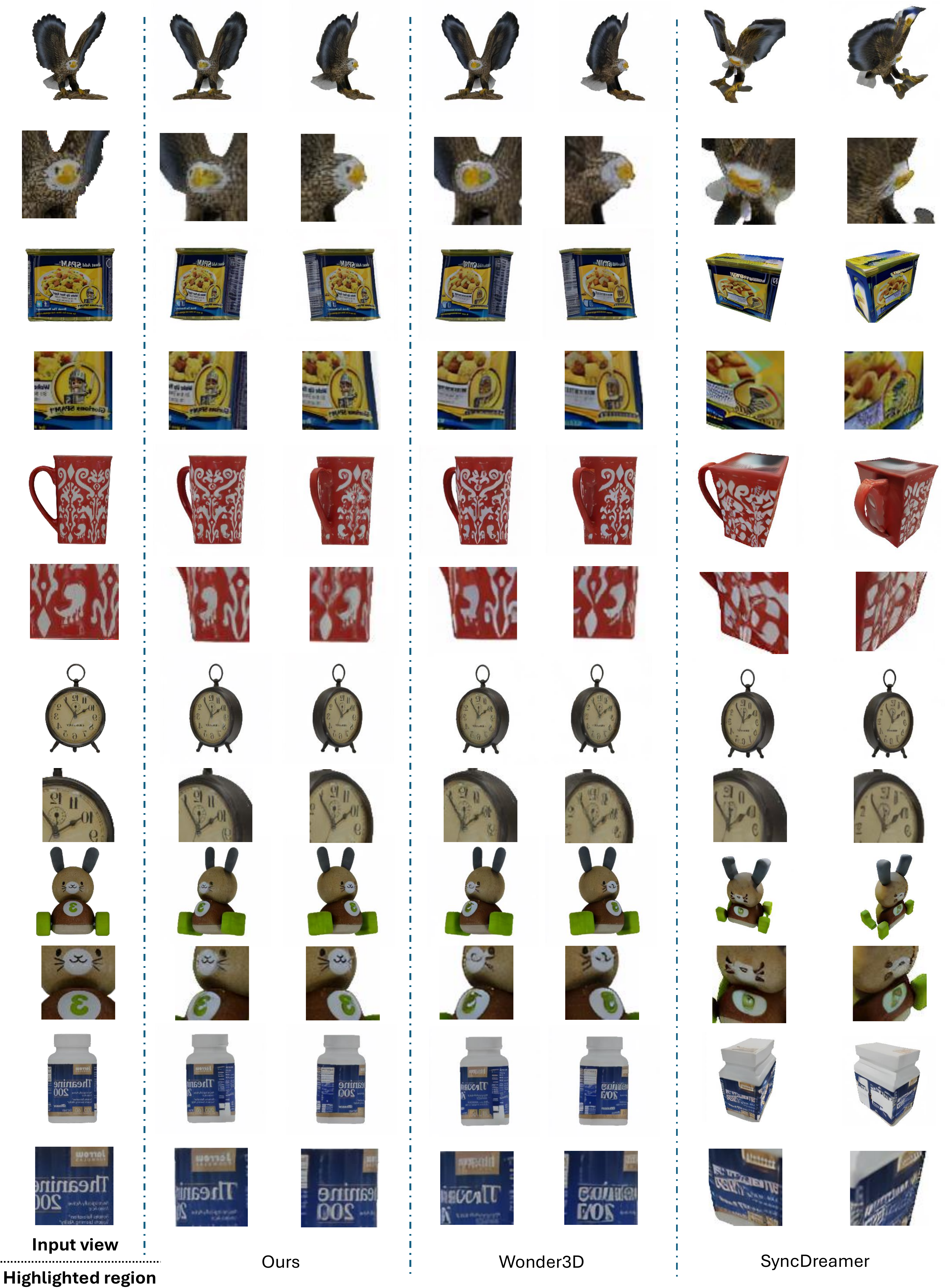}
\caption{
\textbf{Qualitative comparisons with baselines.} 
}
\label{fig:full}
\end{figure*}

\subsection{Implementation details}
\label{sec:imp}
We train our model on a subset of the Objaverse~\cite{objaverse} dataset, which includes around 23k objects with high-quality geometry and texture. To render the dataset, our procedure is similar to wonder3D~\cite{wonder3d}: we render RGB images from six fixed views---front, front right, right, back, left, and front left. Then the images are encoded by the VAE encoder of Stable Diffusion~\cite{ldm}. During inference, the image latents as the decoder's input are generated by wonder3D~\cite{wonder3d} given the input image. Note that our method is compatible to all other latent multi-view diffusion models. We choose wonder3D due to its SOTA performance. The depth map of all views are calculated from a NeuS~\cite{neus} model, which is optimized in the same way as wonder3D~\cite{wonder3d}, i.e., from six normal and RGB maps decoded from the default decoder of Stable Diffusion~\cite{ldm}.

We base our model on the SVD~\cite{svd} decoder VAE and finetune. We use the SVD decoder rather than a Stable Diffusion~\cite{ldm} decoder as the starting point because  the spatio-temporal CNN in the former  improves the multi-view pixel-level consistency. For the truncated epipolar attention, we sample $7,7,7,2$ points around the noised depth and the range is set to $r=0.1$. For the noise scale, we have $0.1=s_1=3s_2=9s_3=9s_4$. Besides low resolution latents, we also condition our model on the front view of the input resolution, which is necessary to maintain detailed texture. Similar to SPAD~\cite{spad}, we also apply a Plucker layer to help the truncated epipolar attention. We retain the optimizer settings from Snapfusion’s decoder training, which means the learning rate is $1\textrm{e}{-5}$ and the loss is a combination of an MSE loss and a perception loss~\cite{lpips}. During fine-tuning, the resolution of all views are 256 and the batch size is 1, the latents of images are $8\times$ downsampled. We use 8 Nvidia Tesla H100 GPUs to train for 3k iterations, which takes around 1 day.

\section{Experiments}
\label{sec:exp}

\begin{table}[t]
    \centering
    \setlength{\tabcolsep}{5pt}
    \small
    \begin{tabular}{lccc}
        \toprule
        \textbf{Method} & PSNR $\uparrow$ & SSIM $\uparrow$ & LPIPS $\downarrow$ \\
        \midrule
        Realfusion  & 15.26 & 0.722 & 0.283  \\
        Zero123     & 18.93 & 0.779 & 0.166  \\
        SyncDreamer & 20.06 & 0.798 & \textbf{0.146}  \\
        Wonder3D    & 20.55 & 0.845 & 0.166  \\
        Ours        & \textbf{20.74} & \textbf{0.847} & 0.164 \\ 
        \bottomrule
    \end{tabular}
    \caption{\textbf{Image-conditioned novel view synthesis on Google Scanned Objects.} We report PSNR, SSIM, and LPIPS on the generated novel view images of GSO objects.}
    \vspace{-5mm}

    \label{tab:rgb}
\end{table}

We conduct extensive experiments to answer the following questions: a) Can our method generate high-quality multi-view images that are consistent and pixel-aligned with the input image and amongst each other? b) Does better pixel-level consistent multi-view generation improve high-quality image-guided 3D asset generation? c) How much do depth-truncated epipolar attention and structured-noise depth augmentation help improve  consistency?

\subsection{Multi-view consistency}

Following prior works, we evaluate baselines and our method on a subset of the Google Scanned Object (GSO) dataset~\cite{gso}, which includes a variety of objects in common life. The subset matches what is used in SyncDreamer~\cite{syncdreamer} and wonder3D~\cite{wonder3d}, including 30 objects from humans and animals to everyday objects. For each object, we render its front view in a 256 resolution and use it as the input to all methods. Moreover, we use the photometric PSNR, SSIM~\cite{ssim}, and LPIPS~\cite{lpips} as evaluation metrics. The quantitative results are summarized in Tab.~\ref{tab:rgb}.
Note that Wonder3D's performance in our evaluation is lower than  reported in the original paper. We tried our best to re-implement their evaluation. Results still improve upon those of other methods. Our method performs favorably to Wonder3D in PSNR and SSIM. 

Qualitatively, the multi-view images generated by our method are more consistent, as shown in Fig.~\ref{fig:full}. We provide zoomed-in illustrations to highlight complex textures.  
Notably, our method generates textures that are more faithful to the input view, while Wonder3D and SyncDreamer both yield blurred textures. 
This is due to their diffusion process  occurring in  latent space with limited resolution. Moreover, their decoder doesn't consider other views.

Next, we quantitatively and explicitly assess the consistency among images to further showcase the necessity of pixel-level alignment.
We measure the number of correspondences between adjacent views using the off-the-shelf dense matching method AspanFormer~\cite{aspan}.
As shown in Tab.~\ref{tab:con}, the proposed method outperforms Wonder3D by $40\%$. This result highlights the improved pixel-alignment.

 \begin{figure*}[th]
 \centering
  \includegraphics[width=0.9\textwidth]{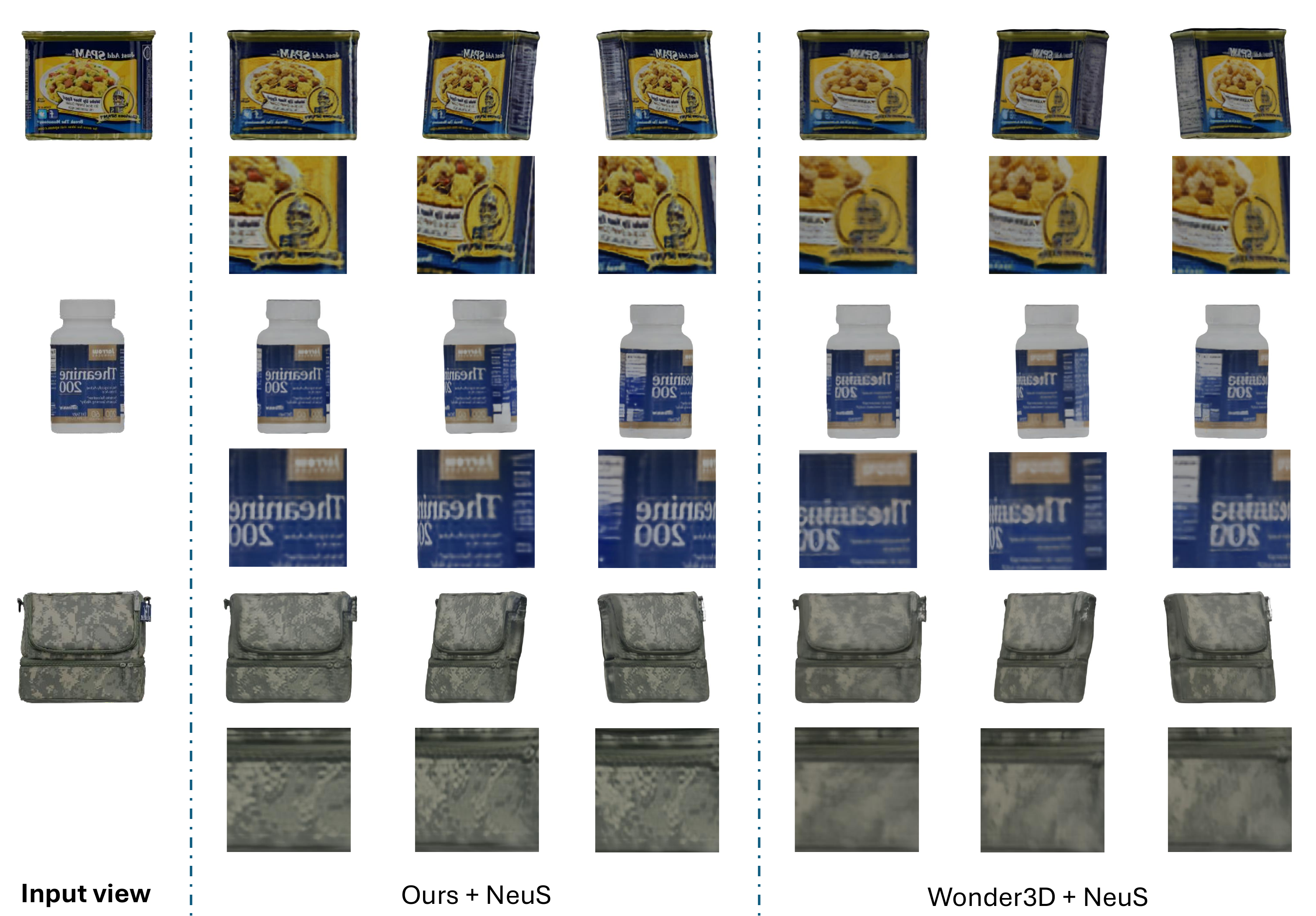}
 \caption{\textbf{Qualitative comparisons after 3D rendering.}
 To better understand the impact of pixel-level aligned multi-view images in the 3D generation pipeline, we reconstruct the 3D object using generated multi-view images. We can clearly observe that  inconsistent multi-view images lead to reconstructed 3D objects which are blurry.
 }
 \label{fig:rec}
 \end{figure*}
 
\subsection{Rerendering from 3D generation}
Next, we show that consistent multi-view generation is beneficial for 3D asset generation. As shown in Fig.~\ref{fig:rec}, we optimize NeuS~\cite{neus} again using the images decoded by our method, and re-render the NeuS results from the fixed views. The  Wonder3D~\cite{wonder3d} baseline reconstruction follows its  procedure. We  observe  the baseline's re-rendering to be much more blurred due to pixel-level misalignment. In contrast, our method's re-rendering remains consistent and maintains the fine details observed in the front view.

\subsection{Ablation study}
\begin{figure*}[h]
 \centering
  \includegraphics[width=0.83\textwidth]{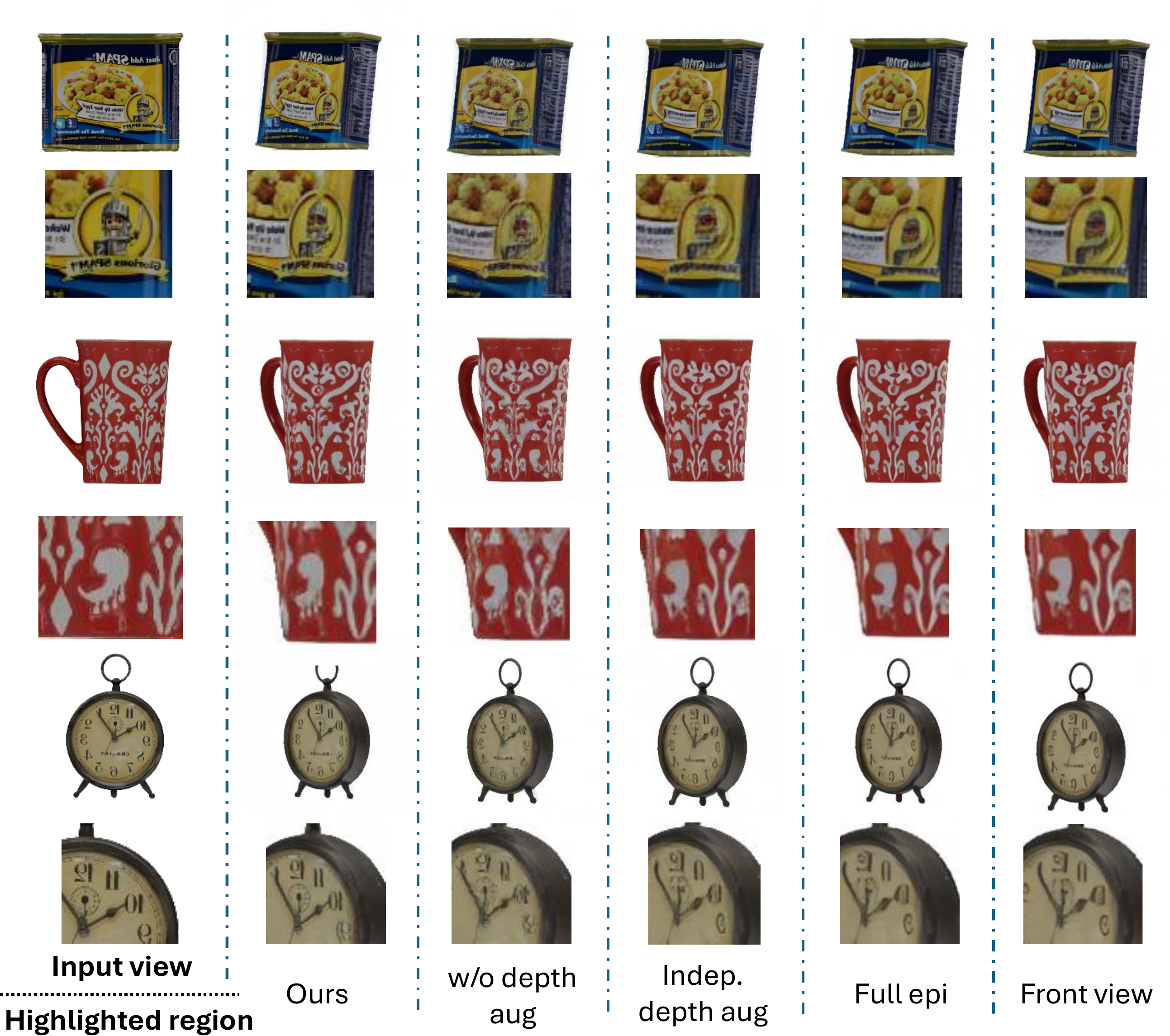}
 \caption{\textbf{Ablation.} We compare with several variants to showcase the necessity of the propose depth-truncated epipolar attention and the structured-noise depth augmentation.}
 \label{fig:abl}
 \end{figure*}

We illustrate how the truncated epipolar attention module and the structured-noise depth augmentation improve consistency. Results are shown in Fig.~\ref{fig:abl}. 
We study several baselines: 
\textit{w/o epi.} only adds the front view condition to the SVD decoder. 
We observe that its output is very blurred. This highlights that learning the consistency via solely the decoder's original CNN and without epipolar attention is difficult. 
Then we assess \textit{Full epi.}, using conventional epipolar attention that queries all adjacent features on the whole epipolar line. Note that due to limited GPU memory, full epipolar attention can  only be applied on resolutions $\le 128$. The results are worse than the truncated version for two reasons. First, excessive irrelevant information is  processed by the attention mechanism, making it difficult for models to focus on the critical information. Second, the attention is not applied to the high resolution, resulting in a less direct impact to the output.

Next, we study the necessity of the structured-noise depth augmentation.
The method abbreviated with \textit{w/o depth aug.} uses ground-truth depth without any augmentation, and the method referred to with \textit{Indep.\ depth aug.} augments data by adding independent noise to each pixel. 
We find that the output of \textit{w/o depth aug.} has severe artifacts due to the large distribution gap between depths used in training and inference. Meanwhile,  \textit{Indep.\ depth aug.} struggles to learn the correct consistency given too much high-frequency noise. These results emphasize that a proper balance is desirable to improve the quality of the results.

Finally, we also report in Tab.~\ref{tab:con} the pixel consistency of all baseline methods. 
Both depth-truncated epipolar attention and the structured-noise depth augmentation play a crucial role in improving consistency. Removing any of our contributions yields a lower number of matching correspondences.

\begin{table*}[h]
\centering
\setlength{\tabcolsep}{4pt}
\begin{tabular}{c|c|c|c|c|c|c|c}
\toprule
Methods     & Ours & Zero123 & Wonder3D & w/o depth aug. & Indep.\ aug. & Full epi. & w/o epi.\\
\midrule
No.\ of corr. & \textbf{458.87} & 54.28 & 329.56 & 291.59 & 259.03 & 254.20 & 245.94 \\
\bottomrule
\end{tabular}
\vspace{2mm}
\caption{\textbf{Evaluating pixel-level alignment.}
To better understand the necessity of pixel-aligned multi-view images, we measure the number of correspondences using AspanFormer.
}
\label{tab:con}
\end{table*}

\section{Conclusion}
\label{sec:conc}

We study pixel-aligned multi-view generation. Multi-view images obtained from multi-view generation are emerging as an important auxiliary representation for 3D generation. However, current works suffer from pixel-level misalignment. We address this issue by improving existing decoders via a depth truncated epipolar attention exploiting depth information, and 
a structured-depth noise augmentation to mitigate the domain gap between depths used in training and inference. We qualitatively and quantitatively show that our method achieves better consistency to the given front view and among predicted cross views. 

\noindent\textbf{Limitations:} 
The back view information is often not clearly shown in the provided front view and only finetuning a reconstruction decoder can not complete the 
texture of unseen views. We leave replacing the decoder with a pixel-level upsampling diffusion model as future work.

\noindent\textbf{Broader impacts.} Content generation in general may have positive and negative societal impacts. On the positive side, generating a desired look, even if abstract, is easier than ever. However, on the negative side, potentially malicious or deceiving content can also be generated easily.

\clearpage

{
    \small
    \bibliographystyle{ieeenat_fullname}
    \bibliography{main}
}

\end{document}